\documentclass{article}

\usepackage{PRIMEarxiv}

\usepackage[utf8]{inputenc} % allow utf-8 input
\usepackage[T1]{fontenc}    % use 8-bit T1 fonts
\usepackage{hyperref}       % hyperlinks
\usepackage{url}            % simple URL typesetting
\usepackage{booktabs}       % professional-quality tables
\usepackage{amsfonts}       % blackboard math symbols
\usepackage{nicefrac}       % compact symbols for 1/2, etc.
\usepackage{microtype}      % microtypography
\usepackage{lipsum}
\usepackage{fancyhdr}       % header
\usepackage{graphicx}       % graphics
\graphicspath{{media/}}     % organize your images and other figures under media/ folder

\usepackage{amsmath}
\usepackage{adjustbox}
\usepackage{multirow}
\usepackage{cleveref}
\DeclareMathOperator{\argmin}{argmin}
\DeclareMathOperator{\aggregator}{aggregator}
\newcommand{\norm}[1]{\lVert #1 \rVert}

\usepackage{xcolor}

\title{Attacks on fairness in Federated Learning}

\author{
  Joseph Rance \\
  University of Cambridge \\
  \texttt{jr879@cam.ac.uk}
   \And
  Filip Svoboda \\
  University of Cambridge \\
  \texttt{fs437@cam.ac.uk}
}

\begin{document}
\maketitle

\begin{abstract}
Federated Learning is an important emerging distributed training paradigm that keeps data private on clients. It is now well understood that by controlling only a small subset of FL clients, it is possible to introduce a backdoor to a federated learning model, in the presence of certain attributes. In this paper, we present a new type of attack that compromises the fairness of the trained model. Fairness is understood to be the attribute-level performance distribution of a trained model. It is particularly salient in domains where, for example, skewed accuracy discrimination between subpopulations could have disastrous consequences. We find that by employing a threat model similar to that of a backdoor attack, an attacker is able to influence the aggregated model to have an unfair performance distribution between any given set of attributes. Furthermore, we find that this attack is possible by controlling only a single client. While combating naturally induced unfairness in FL has previously been discussed in depth, its artificially induced kind has been neglected. We show that defending against attacks on fairness should be a critical consideration in any situation where unfairness in a trained model could benefit a user who participated in its training.
\end{abstract}
\section{Introduction} \label{Intro}

\large{\color{teal} A substantially updated version of this paper can be found under the title ``Can Private Machine Learning Be Fair?". \href{https://ojs.aaai.org/index.php/AAAI/article/download/34216/36371}{\underline{Link}}}

This version of the paper has been extensively summarized to fit the page limit of NeurIPS camera
ready, and some materials, experiments, discussions, and methods are moved to appendix, which
might make some parts hard to follow or cause inconsistencies. To avoid such cases, please read our
arXiv version instead [1]

Federated Learning (FL) has become an attractive machine learning paradigm for training models on private user data \cite{uses}. In FL, each user trains a local model on their data and then sends this model's parameters to a central server where they are aggregated into a single model \cite{fedavg}. This differs from centralised training in that the server does not require any of the users' data. This is an double edged sword, as the lack of transparency in training data then opens the door for a variety of training time backdoor attacks that produce updates which are entirely out of distribution \cite{flbackdoor1, flbackdoor5, flbackdoor6, flbackdoor7, flbackdoor8}. 

Previous backdoor attacks in FL modify the target model to have new functionality in the presence of certain attributes. In this paper we ask - what if the attacker instead wants to modify the existing behaviour to provide them with some relative advantage over the other federation participants? Consider a situation where a group of pharmaceutical companies want to train a shared model to predict drug-target interaction. Perhaps one malicious actor has a vested interest in increasing the strength of this predictive model with respect to a given drug which happens to be their exclusive intellectual property. As all machine learning models are a reflection of their underlying data, a successful attack would, in effect, see the aggregate model reveal more information about the attribute of the data the attacker cares about, than about the others. This effect could not be corrected for by the other clients, as the trained model would not contain the relevant information. And so, if the attacker is successful not only do they obtain unfair advantage with respect to the attribute they care about, but their competitors would also be potentially further hurt by this re-balancing of information content. Finally, later when revealed, this attack could lead to the undermining, and potential unravelling, of any collaboration that led to the federation in the first place.

In this paper we introduce the concept of attacks on fairness in Federated Learning. A similar idea has previously been investigated in the context of centralised models \cite{fairattack1, fairattack2, fairattack3}. However, the threat model in FL is significantly different because we are constrained by controlling a relatively small portion of the clients used to train the FL model. In order for FL systems to become resilient to fairness attacks, a very different set of defences will need to be implemented compared to in the centralised case. We discuss these differences further in \Cref{sec:relfair}.\\

\newpage

Our contributions are:
\begin{itemize}
    \item We introduce the concept of attacking fairness in FL and explain the threat model (\Cref{sec:threat}).
    \item We present and evaluate a new attack on the FL training process that targets model fairness, finding that an attacker is able to influence the aggregated model to become unfair while controlling only a minor subset of clients.\footnote[1]{our code can be found at: \url{https://github.com/slkdfjslkjfd/fl_fairness_attacks}}
    \item We discuss how existing FL backdoor defences could be modified to defend models against fairness attacks (\Cref{sec:defence}).
\end{itemize}

Section \ref{Methodology} introduces our novel threat model and the theoretical underpinning of the fairness attack, as well as an intuitive explanation of it in figure \ref{fig:vectors}. Section \ref{Eval} presents the experimental results supporting our findings. Section \ref{RelWork} reviews the related work that can serve as a useful reference in contextualizing and further building on what the preceding two sections covered. Finally, our conclusions are presented in the final section, section \ref{conclusion}.
\section{Methodology} \label{Methodology}

\subsection{Fairness in Federated Learning}

Previous work has discussed at length the concept of fairness in FL. \cite{multilevelfairness} describes three different types of fairness in FL systems. In this paper we are interested in compromising \textit{attribute level fairness}. That is, our attack aims to influence the model to create a significant discrepancy between the expected accuracy of data points with the targeted attribute and the overall test set accuracy. We discuss further the relationship of our work, to that aimed at the centralized setup in \Cref{sec:relfair}.

\subsection{Threat Model}\label{sec:threat}

Our threat model assumes the attacker knows the current model weights and architecture, which is sent to all clients, as well as an estimate for the number of genuine clients. However, the attacker does not know the model updates sent by these clients. The adversary entirely controls the model updates sent from a non-majority subset of the clients that are used to train the Federated Learning model. Specifically, we assume that fewer than 10\% of clients will maliciously produce model updates. Since the attacker does not control a majority of clients, in order to reduce \textit{attribute level fairness} it is also necessary to compromise \textit{client level fairness}.

\subsection{Attacks on Fairness}

Fairness attacks in the FL setting encounter a new set of challenges. Specifically, while in the centralised case, the main consideration is how to secretly introduce the backdoor into the training procedure, in the federated case the attacker is limited to the production of tainted models. As the malicious models are aggregated with models produced by other clients to produce the server model, the attacker ends up competing with the legitimate clients for the control of the aggregated model.

Furthermore, there is a fundamental difference in the focus of the traditional backdoor attacks, and fairness attacks. Backdoor attacks usually add new functionality in the presence of some trigger that is not usually present in the model's benign dataset. Therefore, these attacks do not conflict with the main task that the FL system is being trained to solve. By contrast, attacks on fairness produce updates that may oppose updates produced by benign clients.

Put together, our attack must therefore modify its update to account for the competing benign clients. If the update we want to make to the aggregated model is $\mathbf{m}$ and we have update $\mathbf{u}_i$ from the $i$th non-malicious client, trained on a dataset of size $n_i$, then the update $\mathbf{v}$ that our malicious client should make is given by

\begin{align}
\underset{\mathbf{v}}{\argmin}\;\norm{\mathbf{m} - \aggregator(\mathbf{v}, \mathbf{u}_0, \mathbf{u}_1, \ldots)}
\end{align}

We will refer to $\mathbf{m}$ as the \textit{target} update, $\mathbf{u}_i$ as the \textit{clean} update, and $\mathbf{v}$ as the \textit{malicious} update.\\

\newpage

\begin{figure}[h]
\centering
\makebox[\textwidth][c]{\includegraphics[scale=0.15]{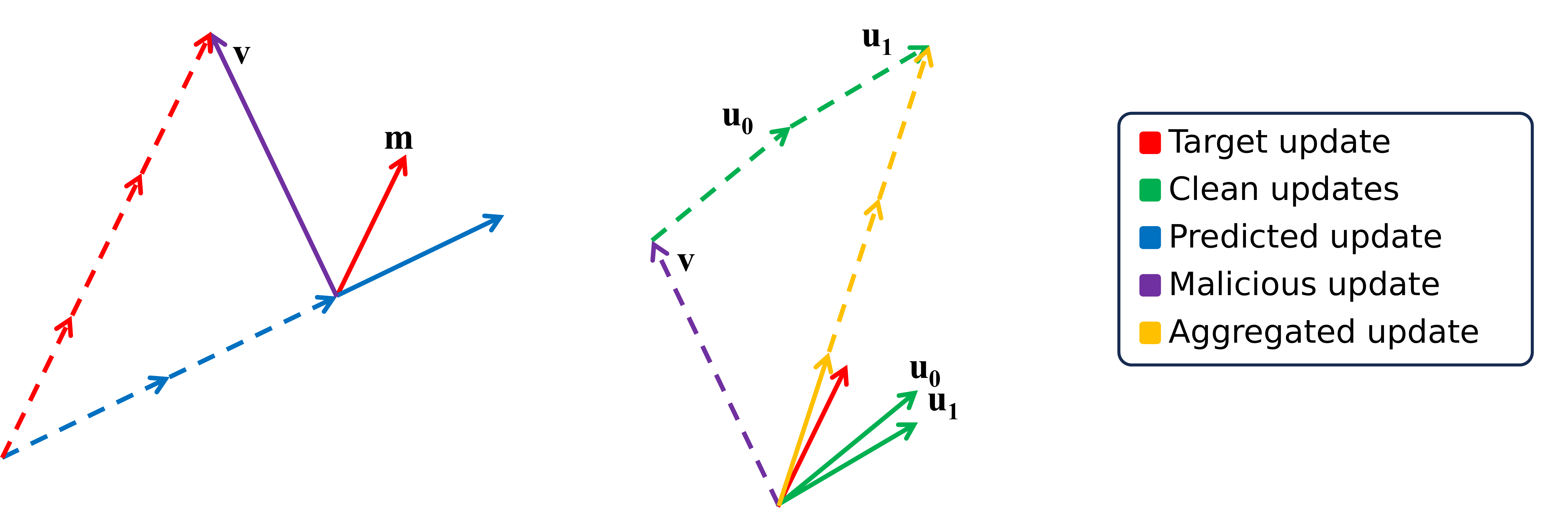}}
\caption{Illustration of how the update vectors aggregate to a value similar to the target. Here, we assume that each client reports the same number of datapoints, $n_i$. The left group of vectors represent the computation done by the malicious client to obtain the target update (purple), while the right group of vectors represent the work done by the aggregator to obtain the aggregated update (yellow). We want to predict the predicted update (blue) to be close to the clean updates (green) so that the aggregated update (yellow) is close to the target update (red)}
\label{fig:vectors}
\end{figure}

If we know the values of each $\mathbf{u}_i$ and the aggregator is sufficiently uncomplicated, we could directly compute the value of $\mathbf{v}$ to produce whichever updates we want. For example, in the case of FedAvg \cite{fedavg}, if we can guess the total number of datapoints trained on in the current round, $n_0\mathbf{v}$ can be computed with:

\begin{align}
\mathbf{m} &= \frac{n_0}{n}\mathbf{v}+\sum^m_{i=1}\frac{n_i}{n}\mathbf{u}_i\\
n_0\mathbf{v} &= n\mathbf{m}-\sum^m_{i=1}n_i\mathbf{u}_i
\end{align}

where we control both $n_0$ and $\mathbf{v}$. In practice, we can only estimate the values of $\mathbf{u}_i$ by predicting updates with our own representative dataset. \Cref{fig:vectors} shows a visual representation of how the FedAvg aggregation will reproduce our target update $\mathbf{m}$.

In order to find $\mathbf{m}$, we simply train our local model with a dataset only consisting of the attributes we want to bias towards. Since this represents a reduced problem, we can expect the accuracy of the target attributes to increase, while the neglected other data may lose accuracy. If we wanted a stronger attack in the case where the model has already been partially trained when our malicious client joins, we could compute an update that unlearns existing behaviour for data not containing the target attributes using an unlearning procedure such as \cite{unlearning}. In our testing, we have found that unlearning was generally not necessary to reduce accuracy on data that does not contain our target attribute.

It is important to note that if the attacker only wanted a model with increased accuracy on specific attributes, then it would be more efficient to simply train a local model on the target updates. Therefore, the attacker doesn't directly benefit from the attack, but rather indirectly by harming the results available to competitors.
\section{Evaluation} \label{Eval}

The goal of our testing is to highlight the effectiveness of our proposed fairness attack on Federated Learning systems. We conduct our experiments on the CIFAR-10 dataset \cite{cifar10dataset}, training a ResNet50 using SGD in a simulated FL environment. Without loss of generality, in our experiments we bias the model towards higher accuracy on classes 0 and 1. We ran our experiments on Nvidia RTX 2080 GPUs.\\

For simplicity, we provided each client with the same amount of clean data, as well as a dataset used by the malicious clients to compute the target update, $\mathbf{m}$, of similar size. All clean datasets were disjoint in order to ensure the malicious client's update prediction was fair. However, the unfair dataset was a subset of the union of the clean datasets. Testing with clients. For simplicity, each client participated in every round and all datasets other than the unfair dataset were i.i.d.. Further testing would be necessary to investigate the effect of higher data heterogeneity and reduced client participation rate.\\

\begin{table}[h!]
\centering
\caption{Percentage accuracies of models trained on all three datasets. Here we report the mean across all classes within each column.}
\label{tab:main}
\adjustbox{scale=0.69}{\begin{tabular}{rccccccccc}
\toprule[0.6pt]
& \multicolumn{3}{c}{3 clients}
& \multicolumn{3}{c}{10 clients}
& \multicolumn{3}{c}{30 clients}\\
Attack
& Target classes
& Other classes
& Overall
& Target classes
& Other classes
& Overall
& Target classes
& Other classes
& Overall\\
\midrule[0.6pt]
Baseline & 92.45 & 90.67 & 91.03 & 94.35 & 90.19 & 91.02 & 91.30 & 85.67 & 86.87\\
\midrule[0.1pt]
Round 80 & 93.35 & 0.34 & 18.94 & 91.60 & 0.73 & 18.90 & 87.05 & 0.11 & 17.50 \\
\midrule[0.1pt]
Full & 89.65 & 0.06 & 17.89 & 78.25 & 0.03 & 15.67 & 84.00 & 0.06 & 16.85\\
\bottomrule[0.6pt]
\end{tabular}}
\end{table}

\Cref{tab:main} presents the results from our testing. We tested two configurations of our attack. The bottom row has the attack performed from the very first training round, while the middle row trained the model on only clean clients for the first 80 rounds. The table shows that our attack has been successful at producing a difference in accuracy between the two targeted classes and the others. Both attack configurations were successful, showing that our attack would work on an existing system, where it was not introduced from the first training round.

\begin{figure}[h]
\centering
\makebox[\textwidth][c]{\includegraphics[scale=0.30]{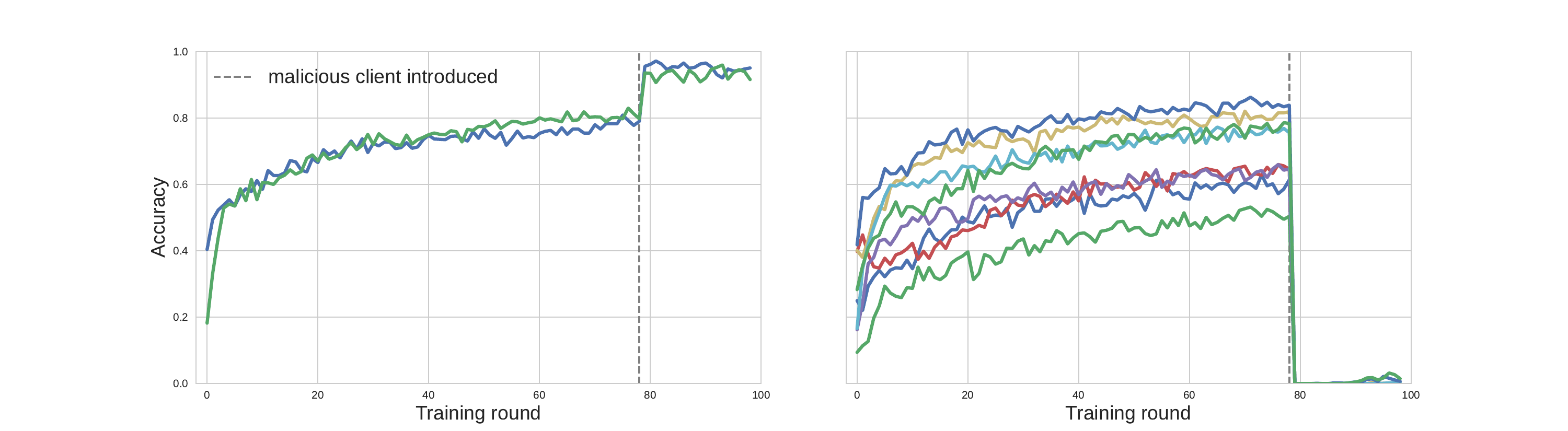}}
\makebox[\textwidth][c]{\vspace{20pt}\includegraphics[scale=0.30]{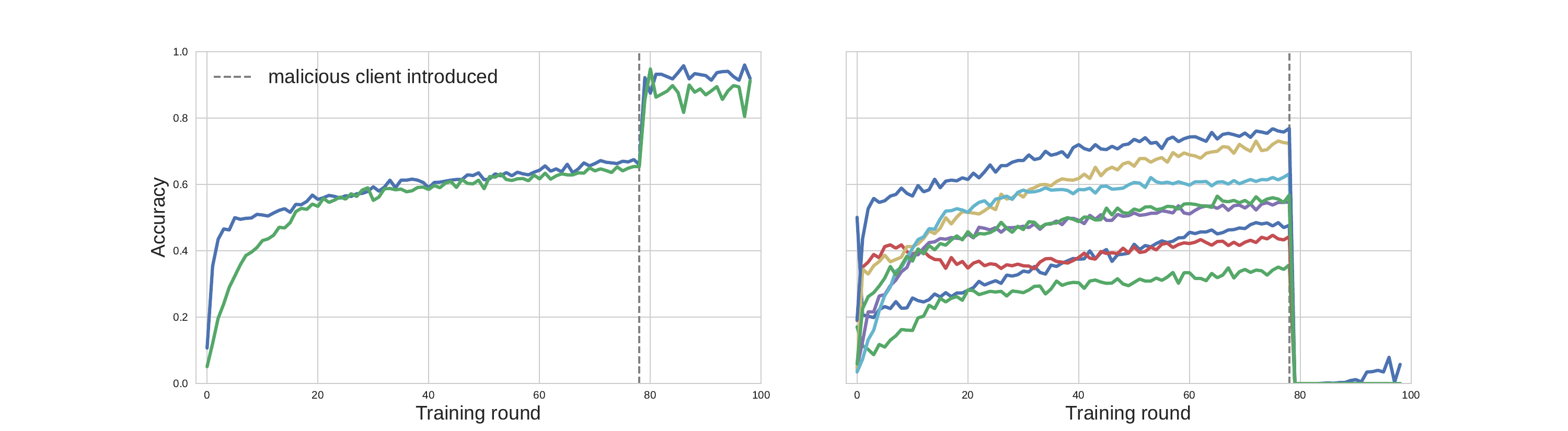}}
\makebox[\textwidth][c]{\vspace{20pt}\includegraphics[scale=0.30]{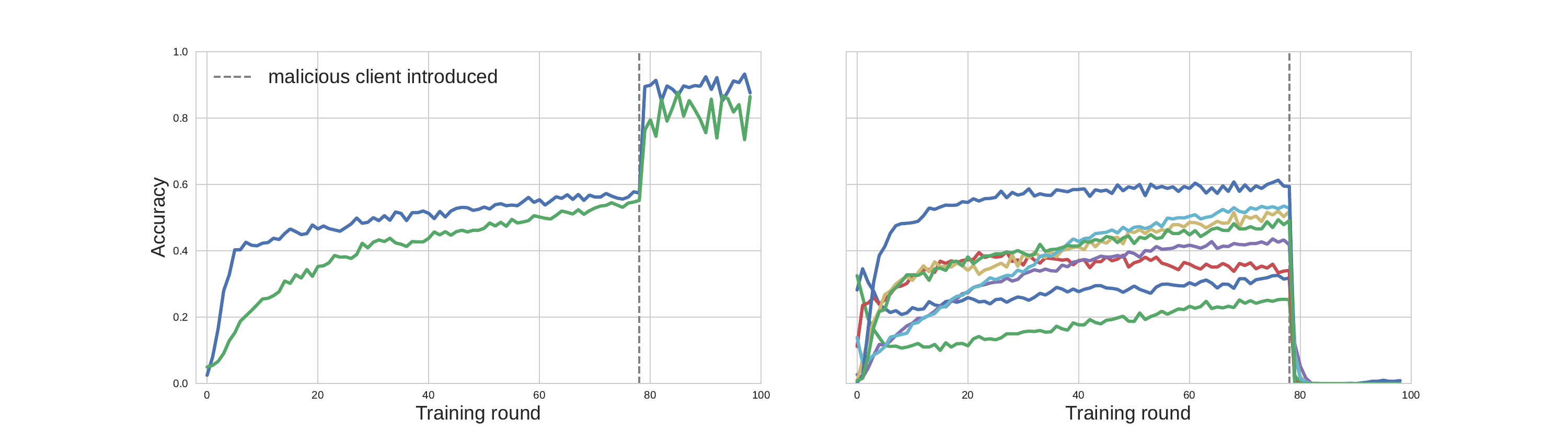}}
\caption{Accuracy of each class per training round of the aggregated model. The left column shows the accuracies for the 2 target classes, while the right shows accuracies for the 8 other classes. The rows show the 3, 10, and 30 client cases from top to bottom. The attack was started on round 80 and produces a clear accuracy discrepancy between its target classes of 0 and 1 and the other classes.}
\label{fig:acctimegraph}
\end{figure}

\Cref{fig:acctimegraph} shows the accuracy of each class per round on the UCI census dataset. We can see that after the attack began on round 10, the target class accuracies immediately becomes higher than that of other classes, as expected.

We tested with 3, 10, and 30 clients, where each experiment contained a single malicious client. Both attack scenarios had slightly less clean data available for training than the baseline due to the data used by the malicious client.

Our results demonstrate that the proposed attack can cause significant accuracy imbalance between subpopulations of our dataset by controlling only a single client. We observe improved performance in the 3 client case than the 10 client and 30 client cases, because each client received more data in this case, making update prediction more reliable.

We also observe a drop in accuracy on the target classes for 10 clients and 30 clients. Since the attack is primarily a denial of service on data which does not include the target attribute, this accuracy drop does not significantly degrade the attack's effectiveness.

It may be possible to achieve a more subtle attack by producing target updates using a dataset that is r, rather than only containing classes 0 and 1. Further testing would be necessary to properly investigate this.

Overall we find that:
\begin{itemize}
    \item An attacker can introduce model unfairness by controlling a minor subset of clients in a FL system.
    \item Our attack can be inserted to models which have already learnt the intended behaviour without significant unfairness towards the attributes we are targeting.
\end{itemize}

\section{Related Work} \label{RelWork}

\subsection{Backdoor attacks}

\cite{badnet} introduced the concept of inserting backdoors into models by poisoning the dataset to add new functionality in the presence of a trigger attribute. There have since been many extensions to this idea to insert backdoors through different means. For example, \cite{batchorderbackdoor} simulated an attack similar to that of \cite{badnet} by shaping gradients through a malicious shuffle function, which has been extended to be inserted as part of data augmentation by \cite{augmentationbackdoors}.

In Federated Learning, these data poisoning attacks will still be effective if performed by each client. However, \cite{flbackdoor1} introduced \textit{update poisoning} attacks, where the attacker entirely controls a set of clients, rather than some part of the training pipeline. \cite{flbackdoor4} further investigates model poisoning where the number of malicious clients is severely limited, and \cite{distributedbackdoor} considered distributing the backdoor updates across multiple clients to improve the attack's stealth. Poisoning attacks are difficult to defend against. They can be effective even against many robust aggregation algorithms \cite{flbackdoor3}. \cite{flbackdoor2} shows that if a model is not robust to adversarial examples, it cannot be robust to backdoor attacks.

\subsection{Fairness in Federated Learning}\label{sec:relfair}

Previous work has placed a strong focus on ensuring fairness in Federated Learning in the face of updates produced by clients with heterogeneous data distributions, both for attribute level fairness \cite{flfair1}, and client level fairness \cite{flfair2}. However, all of these techniques have assumed that the data used by clients still exists within the model's intended domain. Our attacks violate this assumption, potentially compromising the efficacy of these methods.

Attacks on fairness in centralised models have been proposed before \cite{fairattack1, fairattack2, fairattack3}. However, the difficulties faced by fairness attacks in the centralised setting are very different to in FL. In centralised attacks, the updates, the attacker can produce, are constrained by how much of the training pipeline they control. In the federated setting, the attacker can control the entire pipeline, but only for a small set of clients, so the challenge comes from creating a disproportionate impact from these clients.

\subsection{Backdoor defences}\label{sec:defence}

There are many existing defences against backdoors in Federated Learning. For example, many defences attempt to remove updates that deviate from the average update received \cite{flbackdoor4, fldefence1, fldefence2, fldefence3, fldefence4, fldefence5}. However, this directly opposes the intention of improving fairness, as updates which come from datapoints with unusual attributes are most likely to mistakenly identified as malicious. Differential privacy has also been suggested as a possible solution \cite{differentialprivacy1, differentialprivacy2}. \cite{fldefence6} introduce FLAME, which attempts to improve the drawbacks of these approaches by using a combination of techniques. Many defences have been proposed that attempt to detect malicious updates fairly \cite{fldefence7, fldefence8}, for example \cite{fldefence9} use redundant gradients to detect clients which are malicious.

Since our attack is similar to backdoor attacks in Federated Learning in that it produces updates that are distributed differently to what is expected from a legitimate client, many of these defences are likely to have similar effectiveness against our attack without any modification. However, methods for bypassing defences such as \cite{flbackdoor3} may also transfer to our attack. Further experimentation would be needed to test these claims.

Additionally, assuming all clean and target updates have relatively similar direction and magnitude $x$, Equation 3 implies that the minimum length of the malicious update, when it is aligned in the same direction as the clean updates, will have a magnitude of approximately
\begin{align}
\frac{n-\sum^m_{i=1}n_i}{n_0}x&=x
\end{align}
Since the directions are likely to be quite different, we can expect that this attack would produce updates that are generally larger than $x$, and therefore would at a minimum need to be updated to clip the malicious updates to prevent detection based on magnitude.
\section{Conclusion} \label{conclusion}

We have shown that an attacker is able to influence an FL model to have an unfair distribution of accuracies across different attributes by controlling only a small portion of the clients used in training. This is a significant issue because unfairness in machine learning models has been shown to have dire real world consequences\cite{facediscrimination}. There has previously been significant research into fairness for FL in the presence of heterogeneous data distributions across clients, yet the case that we discuss here has so far seen no discussion. In this paper, we show that defending against attacks on fairness should be a critical consideration in any situation where unfairness in a trained model could benefit a user who participated in its training.\\

Previous work on backdoor defences may be effective against this new attack on fairness. In the future we would like the build on this work to investigate the effects of common backdoor defences on the attack we propose here and whether it is possible to bypass these defences.

%Bibliography
\bibliographystyle{unsrt}
\bibliography{paper}

\end{document}